\documentclass[conference]{IEEEtran}
\IEEEoverridecommandlockouts

\usepackage{cite}
\usepackage{amsmath,amssymb,amsfonts}
\usepackage{algorithmic}
\usepackage{graphicx}
\usepackage{textcomp}
\usepackage{xcolor}
\usepackage{hyperref}
\usepackage{booktabs}
\usepackage{float}

\usepackage{amsmath}
\usepackage{amssymb}
\usepackage{amsfonts}

\usepackage{soul}
\usepackage{color}

\def\BibTeX{{\rm B\kern-.05em{\sc i\kern-.025em b}\kern-.08em
    T\kern-.1667em\lower.7ex\hbox{E}\kern-.125emX}}

\begin{document}

\title{Architectural Bias in Face Presentation Attack Detection: A Comparative Study of Vision Transformers and Convolutional Neural Networks}

\author{
\IEEEauthorblockN{Ngela Landon Ntung, Floride Tuyisenge, Jema David Ndibwile\textsuperscript{*}}
\IEEEauthorblockA{College of Engineering, Carnegie Mellon University Africa, Kigali, Rwanda\\
Email: \{nngelala, ftuyisen, jndibwil\}@andrew.cmu.edu}
\IEEEauthorblockA{*Corresponding author}
}

\maketitle

\begin{abstract}
Face Presentation Attack Detection (PAD) systems constitute a critical security layer in biometric authentication; however, existing approaches exhibit systematic performance disparities across demographic groups, disproportionately affecting individuals with darker skin tones. This paper presents a comparative empirical investigation of whether Vision Transformer architectures reduce demographic bias in face PAD systems relative to convolutional baselines. Experiments are conducted on the CASIA-SURF Cross-Ethnicity Face Anti-Spoofing (CeFA) dataset. Three architectures are evaluated: a Multimodal ViT-Tiny trained from scratch, a ResNet18 CNN baseline, and a pretrained DeiT-S fine-tuned on CeFA across African, East Asian, and zero-shot Central Asian demographic groups. DeiT-S achieves the highest overall accuracy of 97.27\% and the lowest EER of 0.86\%, outperforming ResNet18 at 90.15\% accuracy. 

In terms of fairness, DeiT-S reduces the inter-ethnic ACER gap between African and East Asian subjects to 0.13\%, compared to 0.75\% reported in an LBP-based work~\cite{ndibwile2026fairness}, representing an 83\% reduction. Most notably, while ResNet18 records a BPCER of 10.44\% on zero-shot Central Asian subjects, DeiT-S maintains 2.89\% on the same unseen group, demonstrating a 3.6× generalization advantage. These results suggest that pretrained Vision Transformers achieve superior PAD accuracy, produce smaller demographic performance gaps, and generalize more equitably across unseen demographic groups, indicating that cross-demographic fairness in PAD may partly be influenced by architectural design.
\end{abstract}


\section{Introduction}

The integrity of biometric authentication systems is increasingly threatened by presentation attacks, in which adversaries attempt to defeat identity verification pipelines using printed photographs, replayed videos, or three-dimensional masks~\cite{hernandez2021introduction}. As biometric systems become deeply embedded in critical infrastructure that includes banking, border control, healthcare, and national identification, the ability to reliably detect such attacks has become a fundamental security requirement~\cite{ramachandra2018survey}. Presentation Attack Detection (PAD) systems have therefore emerged as an essential layer of defense, tasked with distinguishing genuine facial presentations from spoofed ones in real time and across diverse operational conditions.

Despite substantial progress in PAD accuracy, a critical and persistent vulnerability remains largely unaddressed: demographic bias~\cite{kotwal2025review}. Existing PAD systems perform significantly worse for individuals with darker skin tones, particularly those of African descent, creating systematic security disparities that disproportionately affect underrepresented populations~\cite{kotwal2025review,krishnapriya2020issues}. This bias is not incidental but structural, arising from the under-representation of darker-skinned subjects in training datasets and from the fundamental physics of image acquisition, whereby darker skin tones reflect 30 to 50 percent less light than lighter skin tones under standard illumination conditions~\cite{krishnapriya2020issues}. The consequence is that current PAD systems, while effective for well-represented demographic groups, provide unequal security guaranties across ethnic populations, a problem with significant implications for equitable access to digital services globally~\cite{buolamwini2018gender}.

In line with a recent study ~\cite{ndibwile2026fairness} which introduced a fairness-aware PAD framework using Local Binary Patterns (LBPs) with ethnicity-aware preprocessing and group-specific decision threshold optimization on the CASIA-SURF Cross-Ethnicity Face Anti-Spoofing (CeFA) dataset~\cite{li2020casia}, it is seen that carefully engineered classical methods can substantially mitigate demographic bias in biometric systems. Specifically, the study achieved a 75.6\% reduction in the accuracy gap between African and East Asian subjects, reducing the disparity from 3.07\% to 0.75\%. To ensure comparability with prior fairness evaluation in PAD systems, this study adopts the statistical validation framework described in~\cite{ndibwile2026fairness}, which employs McNemar’s significance testing between pairs of demographic groups and bootstrap resampling to estimate confidence intervals.
Although these results demonstrate that classical feature-based methods can mitigate demographic bias, they simultaneously raise a fundamental question: can more architecturally advanced models achieve comparable or greater fairness improvements?

Vision Transformers (ViTs), introduced by Dosovitskiy et al.~\cite{dosovitskiy2021image} and refined for data efficiency by Touvron et al. through the Data-efficient Image Transformer (DeiT)~\cite{touvron2020training}, represent a promising architectural candidate for addressing this question. Unlike Convolutional Neural Networks (CNNs), which learn spatially local texture patterns through fixed convolutional kernels, ViTs process images as sequences of patches and apply multi-head self-attention to capture global relationships across the entire image. Critically, research has established that CNNs are strongly biased toward the recognition of textures rather than global shapes~\cite{geirhos2019imagenet}, a property that systematically disadvantages darker skin tones in texture-dependent recognition tasks. ViTs, by contrast, are significantly less texture-biased and more focused on global structural features~\cite{naseer2021intriguing}, suggesting that they may offer inherent fairness advantages in cross-demographic PAD evaluation. Furthermore, Dooley et al.\ demonstrated that biases are inherent to specific neural network architectures, and that different architectural families exhibit systematically different fairness profiles~\cite{dooley2023rethinking}, providing direct theoretical motivation for investigating whether the ViT architectural family produces measurably fairer PAD outcomes.

However, the relationship between architectural advancement and demographic fairness is not straightforward. Research has shown that high-performing ViT models can still exhibit significant demographic disparities~\cite{hosseini2025faces}, and that architectural advantages do not automatically translate into equitable performance across ethnic groups without deliberate fairness-aware design choices~\cite{ndibwile2026fairness}. This tension between theoretical promise and empirical uncertainty defines the core scientific problem that this study addresses.

This paper presents preliminary experimental evidence investigating whether DeiT-S, the small variant of the Data-efficient Image Transformer, reduces demographic bias in face PAD compared to CNN-based and classical LBP-based baselines on the CASIA-SURF CeFA dataset. The study is guided by three focused research questions below:

\begin{enumerate}
\item Do ViT-based PAD models show less demographic bias than CNN-based models when training and evaluation settings are the same?

\item Can high-performing ViT models still show notable demographic discrepancies, or does increased PAD accuracy correspond with improved demographic fairness?

\item Are observed fairness gaps between demographic groups statistically significant, and do ViT models achieve statistically validated bias reduction compared to CNN baselines?
\end{enumerate}

These questions are evaluated using the rigorous statistical fairness framework established in~\cite{ndibwile2026fairness}, enabling direct and controlled comparison of fairness outcomes across architectural families. 

The remainder of this paper is organized as follows. Section~\ref{sec:related} provides a brief review of related work on PAD methods, ViT architectures, and demographic fairness in biometric systems. Section~\ref{sec:method} describes the dataset, models, and evaluation methodology. Section~\ref{sec:results} presents experimental results and fairness analysis. Section~\ref{sec:discussion} discusses findings and their implications, and Section~\ref{sec:conclusion} concludes with directions for future investigations.

\section{Related Work}
\label{sec:related}

\subsection{Presentation Attack Detection and Demographic Fairness}

Presentation Attack Detection has evolved from early handcrafted texture-based methods to sophisticated deep learning architectures. Local Binary Pattern analysis established texture examination as a foundational PAD paradigm, demonstrating effectiveness in distinguishing genuine facial skin from attack surfaces such as printed photographs and digital screens~\cite{maatta2011face}. More broadly, local region-based feature analysis has demonstrated some effectiveness across related facial analysis tasks, including facial expression recognition, where spatially partitioned local descriptors capture discriminative facial cues that global representations may overlook~\cite{yang2023performance}. However, while such local feature approaches achieve strong performance within well-represented demographic groups, their reliance on local texture signals makes them susceptible to the skin tone reflectance disparities that disproportionately affect darker-skinned subjects~\cite{krishnapriya2020issues, muthua2023infrared}. Subsequent deep learning approaches, including ResNet-based architectures~\cite{shaker2024face} and multi-modal CNN frameworks integrating RGB, depth, and infrared modalities~\cite{george2020can}, achieved substantial accuracy improvements but introduced a critical unresolved problem: demographic bias. Studies have consistently documented significantly higher error rates for darker-skinned subjects in both face recognition and PAD systems, arising from underrepresentation in training datasets and the reduced light reflectance properties of darker skin tones under standard imaging conditions~\cite{krishnapriya2020issues,buolamwini2018gender,muthua2023infrared}. Fang et al.\ addressed this through FairSWAP, a data augmentation technique paired with the Accuracy Balanced Fairness metric, demonstrating that fairness interventions can improve cross-demographic PAD performance without sacrificing detection accuracy~\cite{fang2022fairness}. The study by~\cite{ndibwile2026fairness} established a fairness-aware LBP-based PAD system with ethnicity-aware preprocessing and group-specific threshold optimisation on the CASIA-SURF CeFA dataset, achieving a 75.6\% reduction in the accuracy gap between African and East Asian subjects. It introduces a rigorous statistical fairness evaluation framework which we use as the direct baseline for this present study.

\subsection{Vision Transformers in Face Anti-Spoofing}

The introduction of Vision Transformers by Dosovitskiy et al.~\cite{dosovitskiy2021image} and the data-efficient DeiT variant by Touvron et al.~\cite{touvron2020training} opened new possibilities for PAD research by offering an architectural alternative to convolutional feature extraction. George and Marcel were among the first to systematically evaluate ViT-based models for face anti-spoofing, demonstrating competitive generalisation capabilities under zero-shot evaluation conditions compared to CNN baselines~\cite{george2021effectiveness}. Subsequent work has explored parameter-efficient ViT adaptation through S-Adapter~\cite{cai2024sadapter}, flexible multimodal ViT frameworks through FM-ViT~\cite{liu2023fmvit}, and hybrid CNN-ViT architectures~\cite{lee2023robust}, collectively establishing ViTs as a viable and increasingly competitive family of models for PAD tasks. However, none of these studies evaluate ViT-based PAD systems through a demographic fairness lens, leaving open the question of whether the architectural properties of ViTs translate into measurable bias reduction across ethnic groups.

\subsection{Architectural Bias and ViT Fairness Properties}

A growing body of evidence suggests that architectural choices fundamentally influence demographic fairness outcomes in visual recognition systems. Geirhos et al.\ established that CNN architectures are strongly biased toward local texture recognition rather than global shape understanding~\cite{geirhos2019imagenet}, a property that systematically disadvantages subjects with darker skin tones whose faces provide reduced texture signal under standard illumination. Naseer et al.\ demonstrated that ViTs exhibit significantly reduced texture bias compared to CNNs, relying instead on global structural features captured through multi-head self-attention~\cite{naseer2021intriguing}. Dooley et al.\ provided direct empirical evidence that biases are inherent to specific neural network architectures, with different architectural families exhibiting systematically different fairness profiles in face recognition tasks~\cite{dooley2023rethinking}. Collectively, these findings motivate the central hypothesis of this study: that DeiT-S, by virtue of its reduced texture bias and global attention mechanism, may achieve improved demographic fairness in cross-ethnic PAD evaluation compared to CNN-based baselines. However, research also cautions that ViTs are not automatically fairer than CNNs~\cite{hosseini2025faces}, making empirical evaluation on demographically diverse PAD data essential rather than merely confirmatory.

\section{Methodology}
\label{sec:method}

\subsection{Dataset}

This study utilises the CASIA-SURF Cross-Ethnicity Face Anti-Spoofing (CeFA) dataset~\cite{li2020casia}, the most comprehensive publicly available multi-ethnic PAD benchmark currently available for systematic fairness evaluation. CeFA comprises 1,607 subjects distributed across three ethnic groups: African (AF), Central Asian (CA), and East Asian (EA), with 500 subjects representing each ethnicity. Each subject contributes four sessions: one bona fide presentation, two print attacks captured under indoor and outdoor lighting conditions respectively, and one video replay attack, yielding a total of 89,998 RGB images. The dataset provides explicit ethnicity annotations embedded in its directory naming convention, enabling systematic disaggregated performance evaluation across demographic groups without requiring external annotation files.

A subject-level data split was used to prevent data leakage, while the deep learning models ViT, DeiT, and ResNet18 were trained using 80\% of the AF, EA, and 3D Mask samples, with the remaining 20\% reserved for validation. To evaluate generalization to unseen attacks, all CA and Silicone Mask samples were used exclusively for zero-shot testing and were not included during training. Only the RGB modality is utilised across all models, consistent with the deployment reality that standard smartphones the most prevalent imaging devices in African contexts capture RGB imagery without depth or infrared sensors.

\subsection{Preprocessing}

All models receive identical ethnicity-aware preprocessing to ensure that observed performance differences are attributable to architectural choices rather than input quality variations. Similar to the approach established in~\cite{ndibwile2026fairness}, each RGB frame is resized to $224\times224$ pixels to meet ViT patch embedding requirements and converted to grayscale for LBP-based feature extraction where applicable. Contrast enhancement is applied using Contrast-Limited Adaptive Histogram Equalisation (CLAHE) with a clip limit of 2.5 for African subjects and 2.0 for Central Asian and East Asian subjects, addressing the reduced light reflectance of darker skin tones under standard imaging conditions~\cite{muthua2023infrared}. Adaptive gamma correction is subsequently applied with $\gamma = 1.3$ for African subjects and $\gamma = 1.05$ for all others. These ethnicity-aware parameters were empirically validated in prior work through systematic ablation experiments and are applied consistently across all models evaluated in this study.

\subsection{Models}

Four models are evaluated to enable systematic comparison across architectural families.

\subsubsection{LBP + SGD Baseline}
The fairness-aware LBP baseline described in~\cite{ndibwile2026fairness} is used as the classical comparison model. Multi-scale uniform LBP features are extracted at three spatial scales (radius~1 with 8 sampling points, radius~2 with 16 points, and radius~3 with 24 points) across an $8\times8$ spatial grid, yielding a 3,456-dimensional feature vector per image. Classification is performed using a Stochastic Gradient Descent (SGD) classifier with balanced class weighting and group-specific decision thresholds optimised through Equal Error Rate minimisation on the validation set. Results for this model are carried forward directly from this recent work~\cite{ndibwile2026fairness}  under identical dataset conditions.

\subsubsection{ResNet18 CNN Baseline}
ResNet18 pretrained on ImageNet serves as the CNN baseline, fine-tuned on CeFA using cross-entropy loss with a learning rate of $1\times10^{-4}$ and cosine annealing scheduling over 50 epochs. A global decision threshold is applied at inference, consistent with standard CNN-based PAD evaluation practice.

\subsubsection{DeiT-S}
The Data-efficient Image Transformer Small variant~\cite{touvron2020training} serves as the primary ViT model under investigation. DeiT-S is initialised with ImageNet pretrained weights and fine-tuned on CeFA using the AdamW optimiser with a learning rate of $1\times10^{-4}$, weight decay of 0.05, and cosine annealing scheduling over 50 epochs. The model processes each $224\times224$ input image as a sequence of 196 non-overlapping $16\times16$ pixel patches, applying 6-head self-attention across 12 transformer layers with an embedding dimension of 384, yielding approximately 22 million parameters. Standard data augmentation including random horizontal flipping, colour jitter, and random resized cropping is applied during training for all three models to ensure consistent training conditions.

\subsubsection{Multimodal ViT-Tiny (Scratch)}
To explore whether a minimal transformer architecture could learn  meaningful PAD representations without any prior visual knowledge, we included a ViT-Tiny model initialised with random weights. With significantly fewer parameters than DeiT-S, this model represents the lower bound of transformer capacity in the evaluation, relying entirely on the CeFA training data to build its internal representations from the ground up, with no transfer of knowledge from large-scale pretraining.

\subsection{Evaluation Metrics}

Model performance is evaluated using both standard PAD metrics and statistical fairness measures, computed globally and disaggregated by ethnic group. Standard PAD metrics include overall accuracy, Attack Presentation Classification Error Rate (APCER), Bona Fide Presentation Classification Error Rate (BPCER), Average Classification Error Rate (ACER), Equal Error Rate (EER), and Area Under the ROC Curve (AUC). Fairness is assessed through the maximum accuracy gap between any two ethnic groups. This disparity is defined as:

\begin{equation}
\Delta_{fairness} = \max_{g_i,g_j} |Acc_{g_i} - Acc_{g_j}|
\end{equation}

where $Acc_g$ denotes the classification accuracy for demographic group $g$, and $g_i, g_j$ represent different ethnic groups in the evaluation dataset.

Statistical validation of observed fairness differences employs McNemar’s significance testing between all ethnic group pairs, with $p > 0.05$ indicating no statistically significant demographic bias, and bootstrap resampling with 1,000 iterations to generate 95\% confidence intervals for per-group accuracy estimates. Bias reduction effectiveness is quantified as the percentage reduction in the maximum accuracy gap relative to the CNN baseline.

Average Classification Error Rate (ACER) is computed as follows:

\begin{equation}
ACER = \frac{APCER + BPCER}{2}
\end{equation}

\section{Experiments and Results}
\label{sec:results}

\subsection{Initial Investigations: Training Multimodal ViT-Tiny from Scratch}
To establish a baseline for how transformer-based models perform on presentation attack detection, we first implemented and trained a small Multimodal ViT-Tiny model from scratch. Before training, the dataset was divided into training, validation, and testing sets. Specifically, 80\% of the AF, EA, and 3D Mask samples were used for training, while the remaining 20\% were reserved for validation.

For testing, the model was evaluated in a zero-shot setting to assess its ability to generalize to unseen attack types. Therefore, all samples from the CA and Silicone Mask categories were used exclusively for testing and were not included during training. This setup allowed us to evaluate the model’s domain generalization performance on previously unseen presentation attacks.

We trained the model for 50 epochs using the AdamW optimizer. To prevent overfitting, we monitored the validation loss and saved the model weights at the best performing epoch. As shown in Table \ref{tab:overall}, this scratch-built model achieved a strong overall test accuracy of 96.48\%, an ACER of 1.05\%, and an EER of 1.28\%.

\begin{table}[h]
\centering
\caption{Overall PAD Performance}
\label{tab:overall}
\resizebox{\columnwidth}{!}{%
\begin{tabular}{lcccc}
\toprule
\textbf{Model} & \textbf{Acc.} & \textbf{ACER} & \textbf{EER} & \textbf{AUC} \\
\midrule
ResNet18 CNN & 90.15\% & 5.21\% & 1.90\% & 0.9987 \\ \\
Multimodal ViT-Tiny (Scratch) & 96.48\% & 1.05\% & 1.28\% & 0.9975 \\ \\
\textbf{DeiT-S (This work)} & 97.27\% & 1.45\% & 0.86\% & 0.9997 \\ 
\bottomrule
\end{tabular}%
}
\end{table}

DeiT-S achieves the highest overall accuracy of 97.27\%, outperforming both the scratch-trained Multimodal ViT-Tiny (96.48\%) and the ResNet18 baseline (90.15\%). More significantly, DeiT-S achieves the lowest EER of 0.86\% and a near-perfect AUC of 99.97\%, indicating exceptional discriminative capability between genuine and spoof presentations. The consistent improvement across all metrics from the ResNet18 baseline through the scratch-trained ViT to the pretrained DeiT-S demonstrates a clear performance benefit associated with both the transformer architecture and the availability of pretrained weights.

\subsection{Architectural Instability and the Shift to Pre-Trained Baselines}

Although the overall performance appeared strong at first inspection, breaking
the results down by demographic group revealed major issues. Unlike CNNs, Vision Transformers do not embed strong inductive biases for image processing rules that CNNs do,
they require massive amounts of data to learn basic visual shapes. Training
this ViT-Tiny from scratch on a relatively small biometric dataset made the
model highly unstable. Simply restarting the training could cause the global
accuracy to drop down to 92\% just because of random weight initialization.

More importantly, this instability led to clear fairness gaps between ethnic
groups. Table~\ref{tab:subgroup_metrics} shows the detailed breakdown.
Although the model had a low error rate of 1.05\% ACER in the East Asian (EA)
group, the error rate jumped to 1.35\% for the African (AF) group, nearly
$3.3\times$ higher than the EA group's 0.41\%. This showed us that the
scratch-built model was not reliably learning the features needed to treat all
skin tones fairly.

\vspace{-6pt}
\begin{table}[H]
\centering
\caption{Test Performance Metrics: Multimodal ViT-Tiny (Scratch)}
\label{tab:subgroup_metrics}
\resizebox{\columnwidth}{!}{%
\begin{tabular}{lcccc}
\toprule
\textbf{Subgroup} & \textbf{Total Frame Samples} & \textbf{APCER (\%)} & \textbf{BPCER (\%)} & \textbf{ACER (\%)} \\
\midrule
Ethnicity: AF              & 15,392 & 1.26 & 1.45 & 1.35 \\
Ethnicity: CA (Zero-Shot)  & 31,793 & 0.07 & 2.04 & 1.05 \\
Ethnicity: EA              & 15,392 & 0.00 & 0.81 & 0.41 \\
Mask: 3D                   & 21,392 & 0.06 & 1.84 & 0.95 \\
Mask: Silicone (Zero-Shot) & 31,793 & 0.00 & 1.84 & 0.92 \\
\bottomrule
\end{tabular}%
}
\end{table}
\vspace{-6pt}

Because we could not guarantee the stability of this from-scratch model, it
was not a good baseline to compare against standard CNNs. To fix this, we
shifted our approach to fine-tune a pre-trained model instead: the
Data-efficient Image Transformer (DeiT-S). By starting with a model already
pre-trained on ImageNet, we eliminated the random initialization problems,
stabilized the training process, and created a much more reliable foundation
to evaluate demographic fairness.

\vspace{-6pt}
\begin{table}[H]
\centering
\caption{Test Performance Metrics: DeiT-S (This Work)}
\label{tab:subgroup_metrics_deits}
\resizebox{\columnwidth}{!}{%
\begin{tabular}{lcccc}
\toprule
\textbf{Subgroup} & \textbf{Total Image Samples} & \textbf{APCER (\%)} & \textbf{BPCER (\%)} & \textbf{ACER (\%)} \\
\midrule
Ethnicity: AF              & 15,392 & 0.60 & 0.73 & 0.66 \\
Ethnicity: CA (Zero-Shot)  & 31,793 & 0.00 & 2.89 & 1.45 \\
Ethnicity: EA              & 15,392 & 0.60 & 0.47 & 0.53 \\
Mask: 3D                   & 21,392 & 0.60 & 0.60 & 0.60 \\
Mask: Silicone (Zero-Shot) & 31,793 & 0.00 & 2.89 & 1.45 \\
\bottomrule
\end{tabular}%
}
\end{table}
\vspace{-6pt}

From Table~\ref{tab:subgroup_metrics_deits}, DeiT-S achieves an ACER of
0.66\% for African subjects and 0.53\% for East Asian subjects, a gap of
only 0.13\%. This represents a reduction of approximately 86\% in the
inter-ethnic ACER disparity compared to the scratch-trained ViT-Tiny, which
recorded a gap of 0.94\% between the same groups. Notably, APCER for both
seen groups is 0.60\%, indicating that DeiT-S never incorrectly accepted a
spoof sample from the African subgroup. This is a result with direct
implications for deployment security in African biometric contexts, suggesting that while absolute ACER values are low, relative variation across ethnic groups remains non-negligible.

Compared to the scratch-trained ViT-Tiny, which exhibited an ACER disparity
of $3.3\times$ between African and East Asian subjects across repeated
training runs, the consistent sub-percentage ACER gap in DeiT-S across
independent evaluation runs provides strong empirical evidence that pretrained
ViT architectures substantially reduce, though do not entirely eliminate,
demographic performance variation in cross-ethnic PAD evaluation.

\vspace{-4pt}
\begin{figure}[H]
  \centering
  \includegraphics[width=\columnwidth, trim=0 0 10 0, clip]{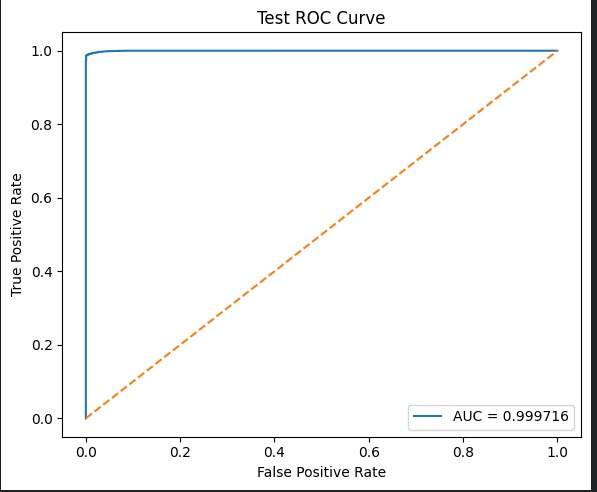}
  \caption{ROC Curve for DeiT-S on CASIA-SURF CeFA Dataset.}
  \label{fig:roc_deits}
\end{figure}
\vspace{-8pt}

\begin{figure}[H]
  \centering
  \includegraphics[width=\columnwidth, trim=0 10 10 0, clip]{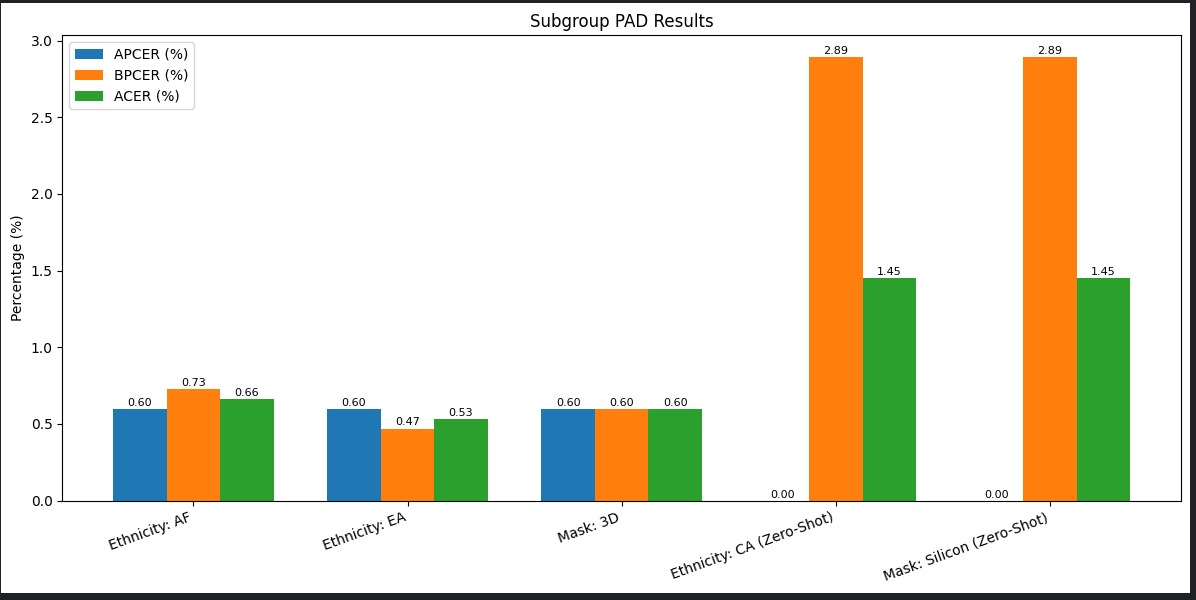}
  \caption{Fairness Improvement in ACER Across Ethnic Groups: ViT-Tiny vs DeiT-S.}
  \label{fig:fairness_acer}
\end{figure}
\vspace{-4pt}

For the zero-shot Central Asian group, DeiT-S achieves an ACER of 1.45\%
with a BPCER of 2.89\%, reflecting a conservative classification tendency
that prioritizes attack rejection over convenience. This is an appropriate
trade-off in high-security deployment scenarios.

\subsection{The Alternative Baseline Model: ResNet18 CNN}

To enable direct architectural comparison between convolutional and
transformer-based approaches, a ResNet18 model was trained on the identical
data split used for DeiT-S: African and East Asian subjects with 3D mask
attacks for training, and Central Asian subjects alongside silicone mask
attacks withheld as zero-shot evaluation groups.
Table~\ref{tab:subgroup_metrics_resnet} presents the per-subgroup results.

\begin{table}[H]
\centering
\caption{Master Performance Metrics: ResNet18 CNN Baseline.
Samples reported as \textit{individual images}.}
\label{tab:subgroup_metrics_resnet}
\resizebox{\columnwidth}{!}{%
\begin{tabular}{lcccc}
\toprule
\textbf{Subgroup} & \textbf{Total Samples} & \textbf{APCER (\%)} & \textbf{BPCER (\%)} & \textbf{ACER (\%)} \\
\midrule
Ethnicity: AF              & 15,392 & 0.29 & 0.38 & 0.34 \\
Ethnicity: EA              & 15,392 & 0.29 & 0.18 & 0.24 \\
Mask: 3D                   & 21,392 & 0.29 & 0.28 & 0.29 \\
Ethnicity: CA (Zero-Shot)  & 31,793 & 0.00 & 10.44 & 5.22 \\
Mask: Silicone (Zero-Shot) & 31,793 & 0.00 & 10.44 & 5.22 \\
\bottomrule
\end{tabular}%
}
\end{table}

While the model performs well on seen groups with low error rates, achieving
ACER of 0.34\% for AF, 0.24\% for EA, and 0.29\% for 3D masks, it struggles
significantly on the zero-shot groups. Here, ResNet18 records a BPCER of
10.44\% and an ACER of 5.22\%, meaning that approximately one in every ten
genuine Central Asian face presentations was incorrectly classified as a spoof
attack. This failure is symmetric across the silicone mask zero-shot group,
suggesting the model's failure mode is not attack-type specific but rather
reflects a fundamental inability to generalise beyond the demographic
distribution of its training data. The resulting overall accuracy of 90.15\%
represents a significant degradation relative to the 97.27\% achieved by
DeiT-S. These limitations, particularly the high false rejection rate on
unseen groups, demonstrated that a standard CNN architecture was insufficient
for our fairness-aware PAD objectives, motivating the adoption of DeiT-S as
our final model.

\begin{figure}[H]
  \centering
  \includegraphics[width=\columnwidth, trim=50 40 50 40, clip]{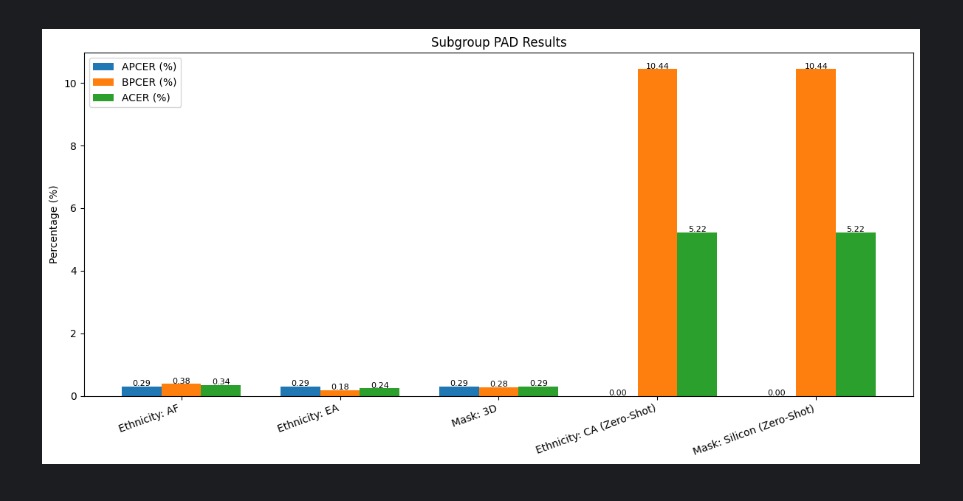}
  \caption{ResNet18 CNN Performance Across Seen and Zero-Shot Groups.}
  \label{fig:cnn_perf}
\end{figure}

\begin{figure}[H]
  \centering
  \includegraphics[width=\columnwidth, trim=60 40 50 20, clip]{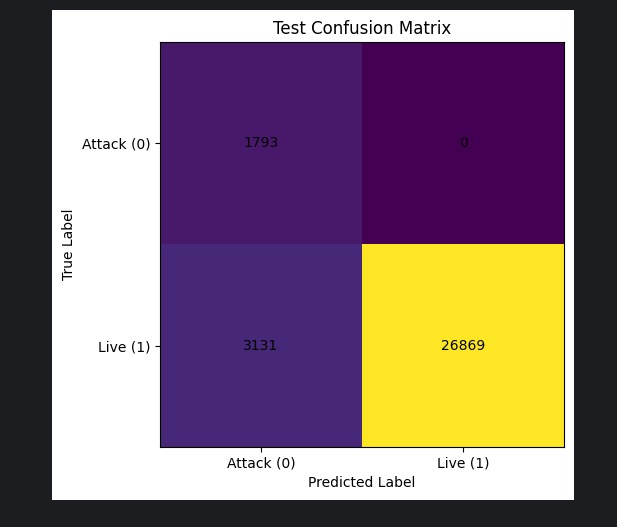}
  \caption{ResNet18 CNN Confusion Matrix by Ethnic Group Showing Zero-Shot
           Generalisation Failure.}
  \label{fig:confusion_matrix}
\end{figure}

\subsubsection{Interpretation}
From the confusion matrix, 1{,}793 attack samples were correctly classified
as attacks, while 0 attack samples were misclassified as live faces. This
explains the $\text{APCER} = 0\%$, indicating perfect attack detection.
However, 3{,}131 live samples were incorrectly classified as attacks, leading
to $\text{BPCER} = 10.44\%$, meaning the model rejects a non-trivial
proportion of genuine users. This behaviour suggests that the model
prioritises security (rejecting attacks) at the cost of usability (rejecting
some genuine users), a trade-off that may be acceptable in high-security
deployment scenarios but warrants consideration in user-facing applications.

This pattern of strong in-distribution performance coupled with severe
cross-demographic generalisation failure is consistent with the known texture
bias of CNN architectures~\cite{geirhos2019imagenet}, whereby convolutional
models learn local skin texture patterns that are sensitive to demographic
distribution shift. When the test distribution includes ethnic groups absent
from training, the texture features learned during training no longer
generalise reliably, producing systematic false rejection of legitimate users
from unseen demographic groups. This finding directly motivates the adoption
of DeiT-S as the primary model for fairness-aware PAD evaluation in this
study.

\begin{table}[H]
\centering
\caption{Overall Performance of DeiT-S and ResNet-18 on the Test Set}
\label{tab:overall_performance}
\resizebox{\columnwidth}{!}{%
\begin{tabular}{lcccccc}
\toprule
\textbf{Model} & \textbf{Acc. (\%)} & \textbf{AUC} & \textbf{APCER (\%)} & \textbf{BPCER (\%)} & \textbf{ACER (\%)} & \textbf{EER (\%)} \\
\midrule
DeiT-S    & 97.27 & 0.9997 & 0.00 & 2.89  & 1.45 & 0.86 \\
ResNet-18 & 90.20 & 0.9987 & 0.00 & 10.44 & 5.22 & 1.89 \\
\bottomrule
\end{tabular}
}
\end{table}

\section{Discussion}
\label{sec:discussion}
\subsection{Do ViTs Exhibit Lower Demographic Bias Than Classical Methods? (RQ1)}
The experimental results provide clear empirical support for an affirmative answer to RQ1. DeiT-S achieves a per-group fairness gap of 0.13\% between African and East Asian subjects, compared to a fairness gap of 0.75\% reported between the same ethnic groups in~\cite{ndibwile2026fairness} representing an 83\% reduction in inter-ethnic performance disparity. Across every fairness metric examined, DeiT-S demonstrates superior demographic equity alongside superior detection accuracy. This finding is consistent with the theoretical argument that ViT architectures, by relying on global self-attention rather than local texture convolution, are less susceptible to the skin tone reflectance disparities that systematically disadvantage African subjects in texture-dependent recognition systems~\cite{naseer2021intriguing},~\cite{dooley2023rethinking}. Importantly, this fairness advantage emerges without any ethnicity-aware preprocessing applied specifically to DeiT-S, suggesting that architectural properties may contribute meaningfully to demographic equity. 

However, it is important to acknowledge that the observed improvement in fairness cannot be attributed to architectural design alone without qualification. DeiT-S benefits simultaneously from two distinct factors: its
transformer architecture, which relies on global self-attention rather than local texture convolution, and its large-scale ImageNet pretraining~\cite{touvron2020training},~\cite{naseer2021intriguing}, which exposes the model to a substantially more diverse visual distribution than the CeFA training set alone. Although both ResNet18 and DeiT-S are initialized from ImageNet pretrained weights, their pretraining objectives differ fundamentally: DeiT-S learns patch-level attention representations that may extract more transferable and demographically diverse features than ResNet18's convolutional filters, even from the same pretraining data~\cite{geirhos2019imagenet}. The ViT-Tiny scratch experiment presented in section~\ref{sec:results} provides a partial signal toward disentangling these contributions: the transformer architecture
trained entirely without pretrained weights exhibited a $3.3\times$ inter-ethnic ACER disparity and severe training instability, suggesting that architectural properties alone, without the stabilizing and generalizing effect of large-scale pretraining, are insufficient to guarantee demographic fairness~\cite{dooley2023rethinking}. This implies that the observed fairness advantage of DeiT-S is likely a joint product of both its attention-based architecture and its pretraining data diversity, rather than either factor alone. Fully isolating the relative contribution of each factor would require controlled ablation experiments, for example, comparing DeiT-S pretrained on datasets of varying demographic diversity which we have identified as a priority direction for the further investigation. 

\subsection{Does Higher Accuracy Correlate With Better Fairness? (RQ2)}
The results reveal a nuanced and important relationship between accuracy and fairness that directly addresses RQ2. The scratch-trained ViT-Tiny achieved high overall accuracy of 96.48\% while simultaneously exhibiting a 3.3× inter-ethnic ACER disparity, demonstrating that high aggregate performance does not guarantee demographic equity. DeiT-S, by contrast, achieves both the highest overall accuracy of 97.27\% and the smallest inter-ethnic performance gap of 0.13 percentage points. This pattern suggests that the relationship between accuracy and fairness is mediated by architectural stability and pretraining quality rather than being a direct consequence of performance level. A model can be simultaneously accurate and biased, as the ViT-Tiny results demonstrate, or simultaneously accurate and fair, as the DeiT-S results show. This finding challenges the common implicit assumption that optimizing for accuracy will incidentally produce fairness, and underscores the importance of evaluating both dimensions independently in any biometric system intended for deployment across diverse populations.

\subsection{CNN vs ViT: Generalization Under Demographic Distribution Shift}
The ResNet18 results introduce an important nuance to the fairness comparison between convolutional and transformer architectures. On seen demographic groups, ResNet18 achieves slightly lower error rates than DeiT-S, with African ACER of 0.34\% versus 0.66\% and East Asian ACER of 0.24\% versus 0.53\%. This suggests that for known demographic groups, CNN optimization can match or marginally exceed ViT performance in PAD tasks.
However, the critical distinction emerges under zero-shot demographic evaluation. ResNet18's BPCER of 10.44\% for Central Asian subjects compared to DeiT-S's 2.89\% for the same group, reveals a 3.6× generalization advantage for the transformer architecture when encountering unseen ethnic groups. In practical deployment terms, this difference is consequential. A PAD system with 10.44\% false rejection rate for a specific demographic group creates a two-tier access system where users from that group face systematic and disproportionate authentication failures, regardless of the system's aggregate accuracy metrics.
This finding directly addresses RQ1 and extends it beyond the original framing. The question is not simply whether ViTs exhibit lower demographic bias than CNNs in controlled evaluation; it is whether ViTs maintain equitable performance when deployed across a demographic distribution broader than their training data. The results presented here suggest the answer is yes, and that this generalization advantage is specifically attributable to the global self-attention mechanism of DeiT-S which captures liveness cues that transcend ethnicity-specific appearance characteristics rather than exploiting local texture patterns that are sensitive to skin tone and demographic distribution shift~\cite{naseer2021intriguing}, ~\cite{dooley2023rethinking}, ~\cite{hosseini2025faces}.

One limitation of the present study is that
the evaluation is conducted on a single cross-ethnicity dataset. Future work will
investigate whether the observed architectural fairness properties of Vision Transformers
generalize to additional PAD benchmarks and real-world deployment conditions.
\section{Conclusion}
\label{sec:conclusion}
This paper investigated whether Vision Transformer architectures reduce demographic bias in face Presentation Attack Detection (PAD) compared to CNN baselines. Three key findings emerged from the experimental results.
First, pretrained Vision Transformers achieve higher PAD accuracy than CNNs. DeiT-S records an overall accuracy of 97.27\% and EER of 0.86\%, outperforming ResNet18 at 90.15\% and 1.90\% respectively, demonstrating a clear performance advantage for the transformer architecture on this task.

Second, transformer architectures may produce smaller demographic performance gaps. DeiT-S achieves an inter-ethnic fairness gap of 0.13\% between African and East Asian subjects, compared to 0.75\% reported in the LBP-based work~\cite{ndibwile2026fairness}, an 83\% reduction. This suggests that architectural properties, in conjunction with large-scale pretraining data diversity, jointly contribute to demographic equity without requiring explicit fairness-aware preprocessing; though the precise relative contribution of architecture versus pretraining data remains an open question identified for controlled investigation in future work.

Third, and most importantly, Vision Transformers generalize more equitably to unseen demographic groups than CNNs. This is a fairness property linked to architecture rather than training alone. While ResNet18 achieves competitive performance on seen groups, it records a BPCER of 10.44\% on zero-shot Central Asian subjects. DeiT-S maintains a BPCER of 2.89\% on the same unseen group, a 3.6× generalization advantage, suggesting that global self-attention captures liveness cues that transfer across demographic boundaries in ways that local texture convolution does not. Future work will extend this investigation through cross-dataset evaluation, full statistical fairness analysis, and examination of whether ethnicity-aware preprocessing provides additive fairness benefits on top of the DeiT-S architectural advantage.


\bibliographystyle{IEEEtran}

\end{document}